# TensiStrength: Stress and relaxation magnitude detection for social media texts[1]

Mike Thelwall

**Computer systems need to be able to react to stress in order to perform optimally on some tasks. This article describes TensiStrength, a system to detect the strength of stress and relaxation expressed in social media text messages. TensiStrength uses a lexical approach and a set of rules to detect direct and indirect expressions of stress or relaxation, particularly in the context of transportation. It is slightly more effective than a comparable sentiment analysis program, although their similar performances occur despite differences on almost half of the tweets gathered. The effectiveness of TensiStrength depends on the nature of the tweets classified, with tweets that are rich in stress-related terms being particularly problematic. Although generic machine learning methods can give better performance than TensiStrength overall, they exploit topic-related terms in a way that may be undesirable in practical applications and that may not work as well in more focused contexts. In conclusion, TensiStrength and generic machine learning approaches work well enough to be practical choices for intelligent applications that need to take advantage of stress information, and the decision about which to use depends on the nature of the texts analysed and the purpose of the task.**

## Introduction

Computer systems can access a constant stream of public social web texts and those that are embedded within communication applications may also store private messages. Harnessing this information to detect stress within individuals or specific groups could allow more intelligent decisions to be made in a wide variety of different contexts. For example, Intelligent Transportation Systems (ITS) harness information from traffic sensors, road monitoring cameras, mobile phone GPS signals, and number plate recognition technology in order to support traffic management (e.g., Wen, Lu, Yan, Zhou, Von Deneen, & Shi, 2011; Knorr, Baselt, Schreckenberg, & Mauve, 2012). There is a constant need to improve the range of sources of evidence for these systems. Stress is indirectly taken into account within ITS through predictable knowledge about stressful times, such as rush hour, and stressful journeys, such as travel to sports events, but accidents or random traffic jams cannot be predicted. To exploit the wealth of text information available to computing systems to improve the predictive power of ITS and other systems, a fast effective method is needed to detect expressions of stress within short informal messages.

The task of deducing affective states from text is already partly solved for sentiment. Opinion mining programs can detect the opinions of users towards products and services from their online reviews or comments with a reasonable degree of accuracy in many contexts (Liu, 2012; Pang & Lee, 2008). Some sentiment analysis programs also attempt to detect a range of emotions, although with limited accuracy (Neviarouskaya, Prendinger, & Ishizuka, 2010). Other systems have also focused on stress detection (see below) but is seems that none harness social media text for this task.

This article introduces a new automated system, TensiStrength, to detect the strength of stress and relaxation expressed in short text messages. It uses a lexical approach to detect indicators of stress or its partial opposite, relaxation, combined with additional linguistic rules to accommodate common ways of modifying the meaning of expressions, such as negation. The result is score for each text on a scale of -1 (no stress) to -5 (extremely stressed) and a parallel score for relaxation from 1 (not relaxing) to 5 (completely relaxed or sleepy). The method is evaluated on a corpus of human-coded tweets and compared to a similar sentiment analysis approach and generic machine learning algorithms. Although TensiStrength is designed for general purpose stress/relaxation detection, its lexicon has additional travel-specific terms for two reasons. First, this research is part of a project to develop automated traffic management and information systems. Second, some stressors are context-specific and it is therefore important to assess whether a stress detection system can work well on a narrow topic.

## Psychological stress

*Biological stress* refers to changes within the body, such as heartrate increases, in response to an unusual stimulus or challenge and in order to prepare for a response (Goldstein & Kopin, 2007; see also: Carr & Umberson, 2013). At the physiological level, stress involves the release of a number of chemicals into the bloodstream to prepare for a fight or flight response, including the stress hormone cortisol (Dickerson & Kemeny, 2004), which increases the blood sugar level. This type of stress can be scientifically detected by heart rate measures or tests for cortisol in saliva, but can also be inferred through self-report questionnaires (Arora et al., 2010). In contrast, *psychological stress* is a subjective feeling of pressure or strain as a result of a stressor in the external environment or internal perceptions of an inability to cope with a situation (Jones, Bright, & Clow, 2001). Stress and negative emotions often co-occur, with each able to be a cause and effect of the other (Lazarus, 2006). Nevertheless, since stress is most closely related to fear and anxiety, it might not be directly triggered by the negative emotion of disgust (e.g., Fredrikson & Furmark, 2006). Conversely, positive emotions have been proposed as an antidote to stress as part of a therapeutic regime for long term sufferers (McCraty & Tomasino, 2006) or as part of a coping strategy (Folkman & Moskowitz, 2000; Moskowitz, Shmueli-Blumberg, Acree, & Folkman, 2012). Other popular strategies for treating or managing stress include cognitive behavioural therapy (Granath, Ingvarsson, von Thiele, & Lundberg, 2006).

Stress can be long term, for example as a by-product of long term medical conditions or psychiatric disorders (Hammen, 1991), or very short term, such as in the fraction of a second between first seeing a crocodile and rationalising that it is safely in a cage. Stress can be broken down into two types: negative stress is termed *distress* whereas positive stress is known as *eustress* (Lazarus, 1974). Positive stress is an important component of some activities that can lead to good outcomes (Simmons, & Nelson, 2001; c.f., Le Fevre, Matheny, & Kolt, 2003). Eustress typically occurs for challenges that are not overwhelming and have a positive outcome.

Events, perceptions or experiences that can cause stress are called *stressors*. Stress is not only caused by immediate fears about physical survival or harm but also by threats to social esteem or successful performance of a task (in a social context), especially when the situation is judged to be uncontrollable (Dickerson & Kemeny, 2004). Hence, there are many factors that might trigger stress and so it would be difficult to precisely delineate all possible stressors. The most common general stressors seem to be interpersonal tensions, work, and social networks (Almeida, Wethington, & Kessler, 2002), but in narrower contexts, the main stressors can be very different (e.g., Aylott & Mitchell, 1999; Cupples, Nolan, Augustine, & Kynoch, 1998; Walker, Smith, Garber, & Claar, 2007).

It is useful to distinguish between long term stressors, including divorce, illness, and major work projects, and transient stressors, which are annoyances or factors with a short term influence during daily life. Stressors of both types can be cumulative in the sense that multiple stressors, even of different types, can tend to aggregate the effect of each other (Pearlin, Schieman, Fazio, & Meersman, 2005). Stressors are subjective, however, and the reaction to a stressor partly depends upon personality type (Semmer, 2006) and the person's perceived ability to cope (Compas, Banez, Malcarne, & Worsham, 1991).

### Stress during travel

Some systematic information has been gathered about travel-related stressors. A non-academic survey of a convenience sample of business travellers found that the major causes of travel-related stress were delayed or lost luggage, poor internet access, economy flights unless short, airport delays, and airport overnight stops (Segalla & Ciobanu, 2012). The flights themselves are also a major source of anxiety for some travellers (Bricker, 2005), as is commuting (Novaco, Stokols, & Milanesi, 1990) and motion sickness (Reason, & Brand, 1975), although some people enjoy travelling (Ory & Mokhtarian, 2005). Jet lag can also be particularly stressful for those performing difficult tasks shortly after a journey (Waterhouse, Reilly, & Edwards, 2004), showing that context is important.

During commuting, factors causing delay are also likely to cause stress (Stokols, Novaco, Stokols, & Campbell, 1978) but the level of stress varies greatly between commuters (Novaco, Stokols, & Milanesi, 1990). External factors are influential here, such as whether the commuter is working in a flexitime job (Lucas & Heady, 2002), the length of the commute, the difficulty of the journey, the gender of the commuter, and whether they are the main driver of a car (Cassidy, 1992; Evans & Wener, 2006; Novaco & Collier, 1994). The unpredictability of the journey time is a stressor for railway commuters (Evans, Wener, & Phillips, 2002), as are crowded trains (Cox, Houdmont, & Griffiths, 2006). Driving is the most stressful mode of transport but walking and cycling are the most relaxing (Wener & Evans, 2011; Gatersleben & Uzzell, 2007).

### Detecting stress

A range of non-textual methods to passively detect stress have been developed. For example, stress cues can be detected in the sound waves of people talking (Lu et al., 2012), and this could become a standard tool within smartphones. A more intrusive approach is to use sensors to detect bodily responses like sweating or heart rate in order to detect specific instances of stress during the day (Bakker, Pechenizkiy, & Sidorova, 2011; Sakr, Elhajj, & Huijer, 2010; Sysoev, Kos, Sedlar, & Pogacnik, 2014) or during a commute (Healey & Picard, 2000). Within social media, a neural network approach has been used to detect short term or longer term stress in social media for individuals, based upon a range of factors, including the emotions expressed in their tweets (Lin et al., 2014). Stress may also be passively inferred from mobile phone call activity patterns (Bogomolov, Lepri, Ferron, Pianesi, & Pentland, 2014) and from oxygen levels in exhaled breath, measured at a distance (Chen, Yuen, Richardson, Liu, & She, 2014).

Long term psychological states, such as post-traumatic stress disorder (Harman, Coppersmith, & Dredze. 2014), have also been estimated from social media. It is possible to detect depression from the profiles of users before its medical diagnosis, through an increased uses of negative sentiment and tight online relationships (De Choudhury, Gamon, Counts, & Horvitz, 2013). A practical application of this might be in the early detection of mothers that are at risk of post-natal depression (De Choudhury, Counts, & Horvitz, 2013). The animal equivalent of social media, in the sense of leaving personal outputs in public, is excretion, and faeces can be analysed chemically to detect traces of stress hormones (Miller, Hobbs, & Sousa, 1991).

There have been some attempts to detect stress through language, although these have typically focused on long term health-related stress rather than stressors or short term stress. The program Linguistic Enquiry and Word Count (LIWC) counts the occurrences of a set of categorised terms (e.g., positive emotion words, pronouns) within texts written by individuals in order to identify patterns that will help to detect psychological differences, such as for diagnosing psychiatric conditions (Tausczik & Pennebaker, 2010). This approach found that the use of more negative words by people writing about a traumatic experience predicts a future lack of health improvements (Pennebaker, 1993). Post-traumatic stress disorder has also been identified through categories of LIWC terms that tend to be used more often by self-declared sufferers in Twitter than by other random users (Coppersmith, Harman, & Dredze, 2014). Hence, it is clear that it is reasonable to analyse the text written by people in order to detect long term stress, although there is no equivalent evidence about short term stress.

## Psychological relaxation

In psychology, *relaxation* refers to a state with the absence of tension and anxiety and is often seen as the polar opposite of stress, anxiety or tension (Pagnini, Manzoni, Castelnuovo, & Molinari, 2013). Relaxation therapies are often proposed for anxiety disorders (Pagnini, Manzoni, Castelnuovo, & Molinari, 2013) and stress (e.g., Iglesias, Azzara, Argibay, et al., 2012; Werneburg, Herman, Preston, et al., 2011). For example, workplace stress can be reduced by yoga programmes (Wolever, Bobinet, McCabe, et al., 2012). Other recognised anti-stress relaxation techniques include breathing exercises, muscle relaxation, jaw relaxation, listening to relaxing music, watching a relaxing film or distraction (e.g., from the cause of anxiety) (De Jong & Gamel, 2006). Mindfulness-Based Stress Reduction (MBSR), which typically encourages a focus on present emotions through meditation and yoga, is also popular and successful at reducing stress (Grossman, Niemann, Schmidt, & Walach, 2004). Despite this, attempts to deliberately relax during stressful situations can have the opposite effect (Wegner, Broome, & Blumberg, 1997) and so stress is not always the polar opposite of relaxation and the two can co-exist to some extent.

At the physiological level, the hormone oxytocin causes relaxation. Its release in the body has a mild sedative effect in the sense of a feeling of relaxation and a more relaxed physiological state (Uvnäs Moberg & Petersson, 2006). It also seems to cause prosocial behaviour (Guastella, Mitchell, & Dadds, 2008; Uvnäs Moberg & Petersson, 2006). Oxytocin can be released as a result of a range of physical factors, such as breastfeeding and touch, as well as through a cognitive appraisal that the situation is suitable for relaxation (Uvnäs Moberg & Petersson, 2006). Oxytocin is to some extent part of an anti-stress system (Uvnäs Moberg, 2003).

Sleep involves several opposite actions to stress, including a decreased heartrate, slower and deeper breathing, lower blood pressure, and a much decreased overall readiness for action (Åkerstedt, 2005). Sleep is also opposed to stress in that people with long term stress often find it difficult to sleep (Basta, Chrousos, Vela-Bueno, & Vgontzas, 2007). The need for sleep can also be a reaction after an experience of short term stress, however, as the body chemistry regains its balance after the incidence (Cespuglio, Marinesco, Baubet, Bonnet, & El Kafi, 1995)

## Sentiment analysis

The task of detecting stress and relaxation is similar to the existing task of detecting sentiment in text. Sentiment analysis systems may classify texts into positive/negative categories (Turney, 2002), positive/negative/neutral categories (Pang & Lee, 2004), by strength of positivity and negativity (Thelwall, Buckley, & Paltoglou, 2012), by valence and arousal (Paltoglou & Thelwall, 2013), by overall polarity and sentiment strength in a single scale (Taboada, Brooke, Tofiloski, Voll, & Stede, 2011), or by strength and emotion type (Neviarouskaya, Prendinger, & Ishizuka, 2010). Programs may attempt to classify entire texts or may extract and classify parts of a text, as in aspect-based (Liu, 2012) and concept based (Poria, Cambria, Winterstein, & Huang, 2014) sentiment analysis, or may focus on polarity shifts within a text (Xia, Xu, Yu, Qi, & Cambria, in press).

Most sentiment analysis systems either use a lexicon or machine learning. The lexical strategy harnesses a sentiment words resource, such as SentiWordNet (Esuli & Sebastiani, 2006) or SenticNet (Cambria, Olsher, & Rajagopal, 2014), and uses an algorithm to predict the presence of sentiment in a text from the words that it contains (Taboada, Brooke, Tofiloski, Voll, & Stede, 2011; Thelwall, Buckley, & Paltoglou, 2012). For example, if the term *good* was registered as positive in a lexicon then texts containing this term would be more likely to be classified as positive. Lexical programs usually call upon a set of additional rules to deal with relevant sentence structures, such as for negation. In contrast, the machine learning approach does not start with lexical resources but instead (typically) is fed with a large set of human-classified texts and learns from these the features to be expected from the different classes. This information is then used to estimate the sentiment of new texts (Pang, Lee, & Vaithyanathan, 2002). Two disadvantages of machine learning are that it is usually opaque to the end user and that it tends to work well only on the types of texts that it is trained for (Blitzer, McDonald, & Pereira, 2006).

Although some papers have analysed sentiment in a travel related context, they have focused on opinions about tourist destinations rather than sentiment about journeys (e.g., Gonzalez-Rodriguez, Martinez-Torres, & Toral, 2014).

## The TensiStrength approach

TensiStrength is an adaptation of the sentiment strength detection software SentiStrength (Thelwall, Buckley, Paltoglou, Cai, & Kappas, 2010; Thelwall, Buckley, & Paltoglou, 2012). The tasks of sentiment and stress/relaxation detection are related but are not equivalent, as argued above. From a valence/arousal perspective, stress seems likely to coincide with high arousal and a negative valence, whereas relaxation may coincide with low arousal and a positive valence. This is not an exact match because positive valence and low arousal could reflect unenthusiastic happiness rather than relaxation. Moreover, low arousal and negativity, such as depression, is a known cause of long term stress, and high arousal and positivity, such as excitement, may help to combat the effects of stress or lead to relaxation after the event.

TensiStrength uses a lexical approach with lists of terms related to stress and relaxation. Informed by research discussed above, TensiStrength's terms are not only synonyms for stress, anxiety and frustration but also terms related to anger and negative emotions because stress can be a response to negative events and can cause negative emotions. The list also includes stressors as additional indirect indicators in the belief that if someone describes situations that are likely causes of stress, such as being late, then they may be experiencing stress. TensiStrength also attempts to detect the opposite state to stress, relaxation, through a parallel approach. This uses a list of terms describing or associating with relaxed situations or relaxed states. These terms also include indicators of positive sentiment. These terms are motivated by the belief that stress tends to be negative and relaxation tends to be positive. The relationship is not straightforward, however, due to eustress and due to joking and laughter as a normal reaction to negative stress (Kuiper & Martin, 1998). For this reason, relaxation is monitored separately to stress, although with rules that relate them (e.g., through negation).

TensiStrength's emotion terms were obtained from SentiStrength, which has a list from LIWC (Pennebaker, Mehl, & Niederhoffer, 2003), General Inquirer (Stone, Dunphy, Smith, & Ogilvie, 1966) and manual additions. The manually added stress terms and indicators of stressors and stressful situations were derived from a range of academic and non-academic sources that describe stress in general or stressors associated with travel. For example, one study found types of commute-related stressors to include heavy traffic, frequent braking, traffic jams, congestion, slow average speeds, transport signals, the overall unpleasantness of the experience, and the need for avoidance strategies (Novaco, Stokols, & Milanesi, 1990), as well as unpredictability of the journey time (Evans, Wener, & Phillips, 2002) and so terms related to these issues were added. Particular attention was given to short term stressors rather than long term sources of stress, such as bereavement. TensiStrength also uses a list of relaxation terms both to identify relaxation and to help identify, through negations, indirect expressions of stress (e.g., not relaxed). Each term in both lists has a numerical strength rating from 1 (nothing) to 5 (very strong). These values were initially assigned intuitively (an unsupervised knowledge-rich approach) and then later refined automatically (a supervised knowledge-rich approach - see below).

TensiStrength exploits its lexicon by assigning to each sentence the score of the highest stress term identified and the highest relaxation term identified. Multiple sentence texts are assigned the highest value of any constituent sentence. The following rules modify this basic approach.

- **Spelling correction** deletes repeated letters to form recognised words.
- **Booster words** strengthen or weaken subsequent words.
- I**dioms** override the scores of their constituent words.
- **Negating relaxing words** turns them into stress words.
- **Negating stress words** neutralises them.
- **At least two repeated letters** added to stress or relation words increase their strength by 1. For instance, woooorrried is more stressful than is worried.
- **Emoticons** are interpreted as relaxed, neutral or stressed, as appropriate, corresponding to their sentiment.
- **Exclamation marks** boost the strength of stress or relaxation within a sentence by 1.
- **Repeated punctuation** with one or more exclamation marks boosts the strength of stress or relaxation within a sentence by 1.

A development corpus of texts was used to help assign the term strengths, to identify missing terms and to refine the sentiment term scores. This was a collection of 3,000 stress-related tweets that were manually classified by the author for stress and relaxation using the 1-5 scale system. These tweets were identified by monitoring a set of stress and relaxation keywords (see appendix) over a week and then taking a random sample. See the methods for a description of the makeup of the evaluation corpus, which was similar to the development corpus. The examples below illustrate the TensiStrength approach.

- *Almost home and the train is delayed*: TensiStrength score 1 for relaxation (the minimum score because there nothing relaxing in the text) and -3 for stress (the term delayed has a stress strength score of -3 in the TensiStrength lexicon).
- *Fell asleep and messed my hair up*: TensiStrength score 4 for relaxation (the term *asleep*) and -1 for stress (no stress terms).
- *Never trust a man with a filthy kitchen*: TensiStrength score 1 for relaxation and -2 for stress. *Trust* is a mild relaxation term (2) but is negated by the previous term "never". *Filthy* scores -2 for mild stress.

The automatic refinement system for term strengths (TensiStrength's supervised mode) uses a hill-climbing approach by assessing whether altering the score of any term in the lexicon could improve the overall accuracy on the development set. This process randomly selects a term from the TensiStrength lexicon, increments its weighting and then rejects the change unless it improves the sum of the positive and negative scores (MAD – see below) by at least 2. For rejected changes, it repeats the process after removing 1 from the term weighting. This is repeated for every term in the lexicon in a random order. If, after checking each term in the lexicon no changes have been made then the process terminates. If at least one change was made then the process repeats.

## Research Questions

The objective of this study is to assess the ability of TensiStrength to detect stress and relaxation in social web texts, with a focus on short term issues and travel in particular. Given that sentiment analysis is likely to be close to stress/relaxation analysis, it is logical to compare TensiStrength to the similar sentiment analysis program that it was derived from, SentiStrength. It is also important to compare it with generic machine learning in order to assess whether machine learning is likely to be more accurate. The following research questions therefore drive the study.

1. Can TensiStrength estimate the strength of stress and relaxation within social web texts with a level of accuracy comparable to humans or at least substantially better than chance?
2. Is the supervised version of TensiStrength more accurate than the unsupervised version?
3. Is TensiStrength more accurate than the similar sentiment analysis program SentiStrength?
4. Can a generic machine learning algorithm based on unigrams, bigrams and/or trigrams be more accurate than TensiStrength?

## Methods

TensiStrength was evaluated by applying it to human-coded texts from a purposive corpus of 3066 English language tweets. These tweets were collected using the keywords from a variety of sources, as described below (see the appendix for a complete list of terms), over a 1 month period in July 2015 and then randomly sampled. The extent to which stress and relaxation appear in tweets depends upon the topic of the tweets and the context and so in order to assess the program on different types of tweet, several small corpora were collected and then combined into a large overall corpus. The number of tweets in each one was approximately proportional to the number of tweets collected overall.

- Common short words: tweets matching a list of common terms and probably reasonably representative of typical tweets. Performance on this set estimates overall performance for Twitter.
- Emotion terms: tweets containing one of a list of emotion terms from the SentiStrength lexicon.
- Insults: tweets containing one of a list of insulting terms taken from the SentiStrength lexicon.
- Opinions: tweets containing one of a list of terms expressing an opinion taken from the SentiStrength lexicon.
- Stress terms: tweets containing one of a list of terms indicating stress or relaxation, taken from the TensiStrength lexicon.
- Transport: tweets containing one of a list of terms relating to travel or transport, manually created for this purpose.

Although four of the six collections above use terms derived from SentiStrength or TensiStrength and hence are not ideal for evaluations it is important to have stress and relaxation-rich sources for evaluation purposes and keyword searches are a logical way to create these. The two other sub-corpora serve as a useful cross-check of the ability of the two programs to work when texts do not necessarily contain any of the keywords in their lexicons.

The tweets were coded by three independent coders using the guidelines in the appendix, following standard practice (Wiebe, Wilson, & Cardie, 2005). The coders were guided by the author's codes on a separate pilot sample, and used several rounds of training on the pilot sample to ensure a high level of consistency. The coders were selected for accurate coding and were not directly part of the TensiStrength project. Inter-coder agreement scores were calculated using an inter-coder weighted Krippendorff α (Krippendorff, 2004), with weight equal to the difference between categories in order to take into account the extent to which the coders disagree. The coders were also compared to each other with Pearson correlations and Mean Absolute Deviation (MAD – the arithmetic mean of the absolute value of the differences between coders) for comparisons with TensiStrength, as well as the percentage of texts that all coders agreed upon, an easily understandable figure. Although there are no accepted minimum values for Krippendorff's α, positive values reflect some degree of agreement between coders, with 1 being the theoretical maximum. The figures are higher than for a previous similar exercise (0.574 for positive sentiment; 0.563 for negative sentiment: Thelwall, et al., 2010) and, in conjunction with the Pearson correlation, this suggests that the coders broadly agreed on the results, although the level of agreement was not high enough to suggest a consensus (Tables 1 and 2).

Table 1. Inter-coder agreement α (top), Pearson correlations (middle) and MAD (bottom) for stress strength on the evaluation corpus for coders A, B and C.

| Corpus | Sample size | A vs. B | B vs. C | A vs. C | Overall α | Full agreement |
|---|---|---|---|---|---|---|
| Common short words | 608 | 0.749<br>0.693<br>0.340 | 0.732<br>0.664<br>0.373 | 0.717<br>0.642<br>0.368 | 0.644 | 54.9% |
| Emotion terms | 619 | 0.701<br>0.630<br>0.435 | 0.684<br>0.620<br>0.465 | 0.677<br>0.580<br>0.480 | 0.584 | 45.1% |
| Insults | 180 | 0.624<br>0.604<br>0.489 | 0.725<br>0.691<br>0.411 | 0.663<br>0.550<br>0.367 | 0.565 | 46.7% |
| Opinions | 476 | 0.742<br>0.695<br>0.332 | 0.737<br>0.692<br>0.380 | 0.713<br>0.633<br>0.355 | 0.641 | 51.9% |
| Stress terms | 655 | 0.704<br>0.660<br>0.441 | 0.694<br>0.628<br>0.466 | 0.694<br>0.606<br>0.467 | 0.597 | 42.0% |
| Transport | 528 | 0.765<br>0.736<br>0.360 | 0.697<br>0.639<br>0.396 | 0.716<br>0.648<br>0.449 | 0.637 | 51.9% |
| **All combined** | **3066** | **0.730<br>0.681<br>0.392** | **0.714<br>0.651<br>0.419** | **0.705<br>0.621<br>0.424** | **0.622** | **48.7%** |

**Table 2**. Inter-coder agreement α (top), Pearson correlations (middle) and MAD (bottom) for relaxation strength on the evaluation corpus for coders A, B and C.

| Corpus | Sample size | A vs. B | B vs. C | A vs. C | Overall α | Full agreement |
|---|---|---|---|---|---|---|
| Common short words | 608 | 0.768<br>0.711<br>0.262 | 0.760<br>0.700<br>0.270 | 0.749<br>0.669<br>0.255 | 0.679 | 64.0% |
| Emotion terms | 619 | 0.787<br>0.725<br>0.197 | 0.690<br>0.571<br>0.250 | 0.721<br>0.633<br>0.276 | 0.645 | 66.7% |
| Insults | 180 | 0.802<br>0.731<br>0.150 | 0.738<br>0.637<br>0.217 | 0.712<br>0.629<br>0.189 | 0.668 | 75.0% |
| Opinions | 476 | 0.806<br>0.775<br>0.200 | 0.740<br>0.678<br>0.233 | 0.758<br>0.691<br>0.248 | 0.692 | 68.5% |
| Stress terms | 655 | 0.696<br>0.625<br>0.292 | 0.694<br>0.568<br>0.281 | 0.696<br>0.595<br>0.270 | 0.595 | 62.0% |
| Transport | 528 | 0.772<br>0.743<br>0.246 | 0.725<br>0.654<br>0.294 | 0.712<br>0.652<br>0.275 | 0.650 | 61.7% |
| All combined | 3066 | **0.767**<br>**0.716**<br>**0.236** | **0.725**<br>**0.638**<br>**0.264** | **0.727**<br>**0.647**<br>**0.261** | **0.654** | **65.1%** |

The performance of the supervised version of TensiStrength was evaluated using 10-fold cross validation 30 times, with the average scores across the 30 iterations recorded. This gives a reasonable estimate of the performance of TensiStrength after optimisation of its lexicon term strengths.

Although exact matches (i.e., precision or the percentage of correct scores) and near misses (i.e., precision +/-1 or percentage of scores within 1 of the correct answer) are natural metrics to use to assess the performance of TensiStrength, the Pearson correlation and MAD are better because both take into account the extent to which the prediction is close to the correct score when it is not correct. Of these two, MAD is better because the correlation has a tendency to reward a conservative strategy of predicting low values because it assesses the tendency for the prediction to be in the same direction as the true score. To illustrate this, suppose that there are four texts, with true values and predictions as follows.

True value: 1, prediction: 1
True value: 5, prediction: 5
True value: 5, prediction: 5
True value: 1, prediction: 5

For this set of texts, the correlation is 0.577 and MAD is 1.000. Suppose that all three predictions of 5 come from a single term with strength 5 that is correct 2/3 of the time but matches a neutral homonym 1/3 of the time. Decreasing the strength of this term to 4, 3, or 2 does not change the correlation strength because the prediction is in the right direction. In contrast, each unit reduction increases MAD by 0.250. Hence, MAD is technically preferable to the Pearson correlation in this situation (where there are many low scores) even though the correlation has the practical advantage that it is a general better understood metric within the social sciences.

A range of different machine learning algorithms were used for stress/relaxation strength detection in order to compare with TensiStrength. The features used were all term unigrams, bigrams and trigrams. Punctuation was included as a term, with consecutive punctuation treated as a single term (e.g., emoticons, multiple exclamation marks). Bigrams and trigrams were not allowed to run across sentences. The numbers of unigrams, bigrams and trigrams were also included as features. These can help if longer texts tend to contain more expressions of stress or relaxation. Feature reduction was applied before machine learning using the information gain metric to give feature sets of size 100, 200,… 1000. Feature reduction can improve machine learning performance by allowing the algorithm to focus on a small number of features that are particularly discriminatory. For example, feature reduction might remove unhelpful words and phrases like "is the", "end.", "it" but keep discriminatory entries such as "scared", "that fucking", "#fuming", and ":)".  The following classifiers were used, representing a range of different machine learning styles in case one worked particularly well. Each classifier was applied using their default configurations in Weka 3.6.

- AdaBoost: An adaptive boosting algorithm based on a simple classifier.
- Bayes: Naïve Bayes likelihood-based algorithm using simplistic assumptions about the independence of features in the data.
- Dec. table: Decision table.
- J48 tree: A classification tree.
- JRip rules: Rule-based classifications.
- Logistic: Simple logistic regression.
- SVM: Support Vector Machines using sequential minimal optimization.

Machine learning relies upon a large amount of training data and so its main application was on the entire corpus, which has the most data. It was also applied to the stress terms sub-corpus, which is important as a stress rich context, and to the transport terms sub-corpus, which represent the most topic-focused context.

## Results

The correlations for the unsupervised version of TensiStrength (upper half of the cells in Tables 3 and 4) show that it performs substantially better than random (a correlation of 0) for both stress and relaxation strength detection, although it also performs substantially worse than the trained human coders. It performs better than unsupervised SentiStrength (lower half of the cells in Tables 3 and 4) on the key MAD metric for both stress and relaxation, with the sole exception of stress strength on the stress sub-corpus.

**Table 3.** Performance metrics for a range of different methods compared to human coders on the evaluation corpus for **stress** strength (TensiStrength: top figures; SentiStrength: bottom figures). The best figures in each pair are bold. Unless stated, the evaluation corpus classifications are rounded to the nearest whole number.

| Corpus | Tweets | Corr. | Exact | Within 1 | MAD |
|---|---|---|---|---|---|
| Common short words | 608 | 0.505 | 327 (53.9%) | **555 (91.4%)** | **0.552** |
| | | **0.511** | **328 (54.0%)** | 551 (90.8%) | 0.562 |
| Emotion terms | 619 | 0.419 | **302 (48.9%)** | **552 (89.3%)** | **0.633** |
| | | **0.467** | 274 (44.3%) | 529 (85.6%) | 0.715 |
| Insults | 180 | 0.329 | **97 (54.2%)** | **168 (93.9%)** | **0.531** |
| | | **0.422** | 76 (42.5%) | 155 (86.6%) | 0.726 |
| Opinions | 476 | 0.452 | **271 (57.1%)** | **437 (92.0%)** | **0.518** |
| | | **0.506** | 232 (48.8%) | 424 (89.3%) | 0.636 |
| Stress terms | 655 | 0.377 | 239 (36.5%) | 517 (79.1%) | 0.893 |
| | | **0.385** | **300 (45.9%)** | **571 (87.3%)** | **0.682** |
| Transport | 528 | **0.457** | 273 (51.8%) | **473 (89.8%)** | **0.594** |
| | | 0.451 | **276 (52.4%)** | 464 (88.0%) | 0.605 |
| All combined | 3066 | 0.444 | **1512 (49.3%)** | **2705 (88.3%)** | **0.642** |
| | | **0.471** | 1489 (48.6%) | 2698 (88.0%) | 0.647 |
| All combined (unrounded) | 3066 | 0.465 | **950 (31.0%)** | **2845 (92.8%)** | **0.660** |
| | | **0.495** | 932 (30.4%) | 2837 (92.5%) | 0.664 |

**Table 4.** Performance metrics for a range of different methods compared to human coders on the evaluation corpus for **relaxation** strength (TensiStrength: top figures; SentiStrength: bottom figures). Unless stated, the evaluation corpus classifications are rounded to the nearest whole number.

| Corpus | Tweets | Corr. | Exact | Within 1 | MAD |
|---|---|---|---|---|---|
| Common short words | 608 | 0.433 | **370 (60.9%)** | **568 (93.6%)** | **0.460** |
| | | **0.475** | 332 (54.7%) | 550 (90.6%) | 0.558 |
| Emotion terms | 619 | 0.359 | **402 (65.1%)** | **577 (93.4%)** | **0.427** |
| | | **0.377** | 344 (55.7%) | 554 (89.7%) | 0.560 |
| Insults | 180 | 0.410 | **125 (69.8%)** | **169 (94.4%)** | **0.363** |
| | | **0.435** | 114 (63.7%) | 165 (92.2%) | 0.453 |
| Opinions | 476 | 0.466 | **316 (66.5%)** | **446 (93.9%)** | **0.402** |
| | | **0.521** | 266 (56.0%) | 420 (88.4%) | 0.562 |
| Stress terms | 655 | 0.347 | **379 (58.0%)** | **599 (91.6%)** | **0.515** |
| | | **0.391** | 349 (53.4%) | 583 (89.1%) | 0.592 |
| Transport | 528 | 0.380 | **316 (60.0%)** | **493 (93.5%)** | **0.482** |
| | | **0.456** | 289 (54.8%) | 484 (91.8%) | 0.535 |
| All combined | 3066 | 0.394 | **1913 (62.4%)** | **2857 (93.2%)** | **0.454** |
| | | **0.442** | 1698 (55.4%) | 2761 (90.1%) | 0.556 |
| All combined (unrounded) | 3066 | 0.422 | **1478 (48.2%)** | **2944 (96.0%)** | **0.467** |
| | | **0.460** | 1285 (41.9%) | 2872 (93.7%) | 0.571 |

The key MAD scores for the supervised version of TensiStrength (upper half of the cells in Tables 5 and 6) are better than the corresponding MAD scores for the *un*supervised version of TensiStrength (upper half of the cells in Tables 3 and 4), showing that supervised mode is preferable to unsupervised mode. The performance of supervised TensiStrength (upper half of the cells in Tables 5 and 6) is about the same overall as that of supervised SentiStrength (lower half of the cells in Tables 5 and 6) on the key MAD metric for stress but it is better for relaxation. This shows that supervision narrows the gap between TensiStrength and SentiStrength on the task of stress strength detection.

**Table 5**. Performance metrics for a range of different methods compared to human coders on the evaluation corpus for **stress** strength with automatically optimised term weightings (10-fold cross validations, average of 30) (supervised TensiStrength: top figures; supervised SentiStrength: bottom figures). The best figures in each pair are bold.

| Corpus | Tweets | Corr. | Exact | Within 1 | MAD |
|---|---|---|---|---|---|
| Common short words | 608 | 0.503 | 328.2 (54.0%) | **559.7 (92.1%)** | 0.544 |
| | | **0.509** | **333.3 (54.8%)** | 557.0 (91.6%) | 0.544 |
| Emotion terms | 619 | 0.431 | **307.7 (49.7%)** | **558.6 (90.2%)** | **0.609** |
| | | **0.448** | 299.0 (48.3%) | 543.7 (87.8%) | 0.646 |
| Insults | 180 | 0.388 | **99.5 (55.3%)** | **168.2 (93.4%)** | **0.524** |
| | | **0.458** | 94.7 (52.6%) | 161.9 (90.0%) | 0.585 |
| Opinions | 476 | 0.464 | **275.4 (57.9%)** | **440.0 (92.4%)** | **0.502** |
| | | **0.493** | 238.3 (50.1%) | 435.3 (91.4%) | 0.596 |
| Stress terms | 655 | **0.373** | 277.2 (42.3%) | 548.7 (83.8%) | 0.769 |
| | | 0.371 | **305.9 (46.7%)** | **582.1 (88.9%)** | **0.654** |
| Transport | 528 | **0.444** | 267.7 (50.7%) | **472.8 (89.5%)** | 0.603 |
| | | 0.437 | **276.8 (52.4%)** | 466.8 (88.4%) | **0.601** |
| All combined | 3066 | 0.461 | 1610.7 (52.5%) | **2806.1 (91.5%)** | 0.567 |
| | | **0.464** | **1628.8 (53.1%)** | 2799.8 (91.3%) | **0.562** |

**Table 6**. Performance metrics for a range of different methods compared to human coders on the evaluation corpus for **relaxation** strength with automatically optimised term weightings (10-fold cross validations, average of 30) (supervised TensiStrength: top figures; supervised SentiStrength: bottom figures).

| Corpus | Tweets | Corr. | Exact | Within 1 | MAD |
|---|---|---|---|---|---|
| Common short words | 608 | 0.428 | **386.0 (63.5%)** | **575.2 (94.6%)** | **0.424** |
| | | **0.453** | 365.2 (60.1%) | 573.4 (94.3%) | 0.461 |
| Emotion terms | 619 | 0.332 | **410.2 (66.3%)** | **580.2 (93.7%)** | **0.407** |
| | | **0.378** | 387.5 (62.6%) | 575.5 (93.0%) | 0.452 |
| Insults | 180 | 0.357 | **129.1 (71.7%)** | **171.0 (95.0%)** | **0.338** |
| | | **0.371** | 125.1 (69.5%) | 167.5 (93.1%) | 0.385 |
| Opinions | 476 | 0.461 | **322.0 (67.7%)** | **452.9 (95.1%)** | **0.378** |
| | | **0.522** | 289.8 (60.9%) | 438.7 (92.2%) | 0.474 |
| Stress terms | 655 | **0.386** | **403.3 (61.6%)** | **620.0 (94.7%)** | **0.442** |
| | | 0.369 | 397.1 (60.6%) | 613.3 (93.6%) | 0.467 |
| Transport | 528 | 0.394 | **324.0 (61.4%)** | 502.8 (95.2%) | **0.439** |
| | | **0.448** | 311.1 (58.9%) | 505.2 (95.7%) | 0.456 |
| All combined | 3066 | 0.425 | **2052.2 (66.9%)** | **2951.4 (96.3%)** | **0.371** |
| | | **0.433** | 1968.3 (64.2%) | 2935.5 (95.7%) | 0.404 |

## Generic machine learning

The generic machine learning methods (with the optimal number of features) applied to the full corpus outperform TensiStrength in terms of all of the metrics except precision (exact) and correlation for stress detection (Table 7) and outperform it across all metrics for relaxation strength detection (Table 8). Thus standard machine learning gives better performance than TensiStrength for relaxation detection, and for stress strength detection as long as correlation is not the primary concern. The choice of method has an effect on the results but most methods perform at least as well as TensiStrength overall.

**Table 7**. Performance metrics for a range of different machine learning methods compared to human coders on the evaluation corpus for **stress** strength with automatically optimised term weightings (10-fold cross validations, average of 30) (n=3066). The number of features is reported for which the optimal metric value was obtained.

| Method | Corr. | Feat. | Exact | Feat. | Within 1 | Feat. | MAD | Feat. |
|---|---|---|---|---|---|---|---|---|
| AdaBoost | 0.137 | 500 | 42.2% | 100 | **98.1%** | **100** | 0.598 | 100 |
| Bayes | 0.264 | 800 | 44.4% | 400 | 88.8% | 400 | 0.683 | 400 |
| Dec. table | 0.238 | 600 | 45.8% | 600 | 97.4% | 200 | 0.570 | 600 |
| J48 tree | 0.213 | 400 | 44.6% | 200 | 90.3% | 400 | 0.658 | 400 |
| JRip rules | 0.130 | 300 | 43.3% | 900 | 97.3% | 300 | 0.596 | 900 |
| Logistic | **0.319** | 200 | 48.9% | 100 | 95.9% | 600 | 0.564 | 100 |
| SVM | 0.296 | 100 | **49.3%** | **200** | 97.3% | 100 | **0.538** | **100** |
| *TensiStrength* | 0.444 | - | 49.3% | - | 88.3% | - | 0.642 | - |

**Table 8**. Performance metrics for a range of different machine learning methods compared to human coders on the evaluation corpus for **relaxation** strength with automatically optimised term weightings (10-fold cross validations, average of 30) (n=3066). The number of features is reported for which the optimal metric value was obtained.

| Method | Corr. | Feat. | Exact | Feat. | Within 1 | Feat. | MAD | Feat. |
|---|---|---|---|---|---|---|---|---|
| AdaBoost | 0.149 | 1000 | 66.3% | 1000 | 94.1% | 100 | 0.398 | 1000 |
| Bayes | 0.324 | 100 | 66.5% | 100 | 95.1% | 100 | 0.385 | 100 |
| Dec. table | 0.324 | 600 | 69% | 200 | 95.6% | 100 | 0.358 | 100 |
| J48 tree | 0.294 | 100 | 67.7% | 100 | 95.5% | 100 | 0.369 | 100 |
| JRip rules | 0.275 | 100 | 67.5% | 200 | 95.0% | 100 | 0.376 | 200 |
| Logistic | **0.480** | 200 | **71.5%** | 200 | **97.1%** | 200 | **0.314** | **200** |
| SVM | 0.438 | 300 | 70.5% | 200 | 97.0% | 700 | 0.328 | 300 |
| *TensiStrength* | 0.394 | - | 62.4% | - | 93.2% | - | 0.454 | - |

On the stress terms sub-corpus, the machine learning methods with the optimal numbers of features outperform TensiStrength across all metrics except correlation for stress detection (Table 9) and the best method outperforms it across all metrics for relaxation strength detection (Table 10). The relatively strong performance of machine learning is despite the relatively small amount of training data available (i.e., 90% of 655 tweets).

**Table 9**. Performance metrics for a range of different machine learning methods compared to human coders on the **stress sub-corpus** of the evaluation corpus for **stress** strength with automatically optimised term weightings (10-fold cross validations, average of 30) (n=655). The number of features is reported for which the optimal metric value was obtained.

| Method | Corr. | Feat. | Exact | Feat. | Within 1 | Feat. | MAD | Feat. |
|---|---|---|---|---|---|---|---|---|
| AdaBoost | 0.069 | 100 | 45.8% | 100 | 96.7% | 100 | 0.575 | 100 |
| Bayes | 0.241 | 100 | 47.7% | 100 | 96.6% | 100 | 0.557 | 100 |
| Dec. table | 0.193 | 100 | 47.8% | 200 | 96.7% | 100 | 0.555 | 100 |
| J48 tree | 0.082 | 500 | 45.8% | 300 | **96.8%** | **100** | 0.575 | 300 |
| JRip rules | 0.199 | 100 | 47.7% | 100 | 96.7% | 100 | 0.557 | 100 |
| Logistic | **0.257** | 100 | 48.8% | 100 | 96.2% | 100 | 0.552 | 100 |
| SVM | 0.255 | 100 | **49.1%** | **100** | 96.4% | 100 | **0.548** | **100** |
| *TensiStrength* | 0.377 | - | 36.5% | - | 79.1% | - | 0.893 | - |

**Table 10**. Performance metrics for a range of different machine learning methods compared to human coders on the **stress sub-corpus** of the evaluation corpus for **relaxation** strength with automatically optimised term weightings (10-fold cross validations, average of 30) (n=655). The number of features is reported for which the optimal metric value was obtained.

| Method | Corr. | Feat. | Exact | Feat. | Within 1 | Feat. | MAD | Feat. |
|---|---|---|---|---|---|---|---|---|
| AdaBoost | 0.196 | 200 | 67.3% | 200 | 96.5% | 100 | 0.365 | 200 |
| Bayes | 0.294 | 100 | 67.9% | 100 | 96.7% | 100 | 0.357 | 100 |
| Dec. table | 0.214 | 300 | 68.1% | 100 | 96.7% | 600 | 0.356 | 200 |
| J48 tree | 0.196 | 600 | 67.8% | 600 | 96.8% | 1000 | 0.357 | 600 |
| JRip rules | 0.176 | 100 | 69.1% | 100 | 96.2% | 600 | 0.352 | 100 |
| Logistic | 0.321 | 100 | 70.0% | 100 | 97.1% | 100 | 0.331 | 100 |
| SVM | **0.383** | **100** | **71.1%** | **100** | **98.0%** | 900 | **0.316** | **100** |
| *TensiStrength* | 0.347 | - | 58.0% | - | 91.6% | - | 0.515 | - |

On the transport terms sub-corpus, TensiStrength outperforms all machine learning methods tested with the optimal number of features across all of the metrics for stress strength (Table 11) but the best machine learning methods outperform it across all metrics for relaxation strength detection (Table 12) on the small sub-corpus (528 tweets).

Table 11. Performance metrics for a range of different machine learning methods compared to human coders on the **transport sub-corpus** of the evaluation corpus for **stress** strength with automatically optimised term weightings (10-fold cross validations, average of 30) (n=528). The number of features is reported for which the optimal metric value was obtained.

| Method | Corr. | Feat. | Exact | Feat. | Within 1 | Feat. | MAD | Feat. |
|---|---|---|---|---|---|---|---|---|
| AdaBoost | 0.071 | 100 | 47.6% | 100 | 80.5% | 600 | 0.745 | 100 |
| Bayes | 0.323 | 700 | 48.3% | 400 | **89.2%** | **1000** | 0.639 | 700 |
| Dec. table | -0.003 | 700 | 47.4% | 100 | 80.3% | 700 | 0.748 | 100 |
| J48 tree | 0.168 | 700 | 48.9% | 400 | 85.7% | 500 | 0.693 | 500 |
| JRip rules | 0.128 | 100 | 48.5% | 400 | 82.4% | 500 | 0.718 | 400 |
| Logistic | 0.168 | 700 | 48.2% | 400 | 84.7% | 600 | 0.708 | 700 |
| SVM | **0.269** | **700** | **50.8%** | **600** | 87.8% | 600 | **0.630** | **600** |
| *TensiStrength* | 0.457 | - | 51.8% | - | 89.8% | - | 0.594 | - |

Table 12. Performance metrics for a range of different machine learning methods compared to human coders on the **transport sub-corpus** of the evaluation corpus for **relaxation** strength with automatically optimised term weightings (10-fold cross validations, average of 30) (n=528). The number of features is reported for which the optimal metric value was obtained.

| Method | Corr. | Feat. | Exact | Feat. | Within 1 | Feat. | MAD | Feat. |
|---|---|---|---|---|---|---|---|---|
| AdaBoost | 0.141 | 100 | 62.8% | 100 | 92.5% | 100 | 0.452 | 100 |
| Bayes | 0.320 | 100 | 63.9% | 100 | 94.0% | 100 | 0.425 | 100 |
| Dec. table | 0.267 | 100 | 64.1% | 100 | 93.8% | 300 | 0.428 | 100 |
| J48 tree | 0.320 | 100 | 65.3% | 100 | 94.1% | 700 | 0.411 | 300 |
| JRip rules | 0.285 | 100 | 63.4% | 100 | 93.8% | 100 | 0.433 | 100 |
| Logistic | **0.424** | **100** | **66.8%** | **200** | 95.7% | 100 | **0.383** | **100** |
| SVM | 0.418 | 100 | 66.5% | 100 | **96.9%** | **1000** | 0.384 | 200 |
| *TensiStrength* | 0.380 | - | 60.0% | - | 93.5% | - | 0.482 | - |

## Discussion

**Limitations** Although the results give evidence about the performance of TensiStrength compared to other approaches, an important limitation is the nature of the corpus used for the evaluation. This was a purposive and relatively balanced set of tweets rather than a random sample. It was constructed to include a range of types of sub-corpora but the performance of the program may vary for other collections of texts, such as those on a specific topic or from sources other than Twitter. Four of the six sub-corpora were constructed from terms in the lexicons of TensiStrength or SentiStrength, and so their performances may be different on other corpora that include other stress-related terms. The methods are also limited by the fact that the coders and programs did not have access to any context about the tweet, such as the name of the tweeter and the text of any related tweets. This information would have improved the classification accuracy of humans more than the classification accuracy of automated methods. There may also be machine learning algorithms that could give better results than the selection used here. The machine learning results are likely to overestimate the accuracy possible in practice because the best performance is from 8 different classifiers, each with 10 different feature set sizes and so these 80 variations give scope for significant capitalisation on chance and hence and unfair advantage for comparisons with TensiStrength, which was only tried with two variations (supervised and unsupervised).

**Overall performance** Subject to the above limitations, the results confirm that TensiStrength is able to estimate the strength of stress and relaxation in tweets with a level of accuracy that is substantially above chance on a range of different sets of tweets, including general tweets, stress-rich tweets, emotion-rich tweets, tweets with insults, opinionated tweets, and transport-related tweets. Its performance was weakest for all metrics on the set of tweets matching stress and relaxation terms. This is a particular concern given that this type of tweet is a likely future application of the program. The low performance on this sub-corpus reflects TensiStrength being likely to classify all tweets as having at least moderate levels of either stress or relaxation and so there would be relatively few easy classifications of texts with no matching terms. This absence could well be a major cause of the reduced overall scores. In other words, a collection of stress/relaxation texts may be the most difficult for a stress/relaxation classification system to process because there will be few obviously neutral texts.

**Comparison with human coders** TensiStrength was substantially less consistent with the average of the three carefully selected and trained human coders than they were with each other and so it is not a suitable replacement for manual coders when high quality results are needed and the time is available to wait for human processing. Nevertheless, it may well have comparable performance to less careful and less well trained coders or randomly selected people because the coders were initially recruited with a competitive process that included a task requiring accurate coding and they used several days of training in order to improve their consistency with the stress and relaxation coding. TensiStrength's advantage over the human coders is speed and expense: whilst each human coder took twelve days to classify 3,066 tweets (one day per week for three months), it took TensiStrength (unsupervised) 0.2 seconds on an office personal computer. TensiStrength therefore makes it practical to conduct huge scale real-time analyses of stress and relaxation in short informal English-language texts, such as tweets. The underlying reason for the poorer performance than for trained coders is probably that language is very flexible, creative and subjective and a simple rule-based system will not understand the meaning or context of a text. This is likely to be particularly true in the presence of homonyms and figurative language, such as sarcasm (Kunneman, Liebrecht, van Mulken, & van den Bosch, 2015). TensiStrength may be able to give comparative performance to typical coders, however, who may misunderstand the task or lose focus due to its repetitive nature.

**Comparison with SentiStrength** TensiStrength is more accurate than the sentiment analysis program SentiStrength overall, although the difference is not large and the two have comparable performance for stress strength in supervised mode. It is surprising that TensiStrength is not substantially more accurate than SentiStrength given that it is based upon SentiStrength but with a series of modifications in order to improve its ability to detect stress and relaxation. Their similar performances reflect positive and negative sentiment being close to relaxation and stress but this may not be true for more narrowly focused collections of texts. On the complete collection of tweets initially collected, after removing duplicates, they agree for positivity/relaxation on 73.6% of texts and for negativity/stress on 73.5% of texts (Tables 13, 14). Hence, despite the overall similar scores they disagree on over a quarter of the tweets and for some of the remaining tweets they may give identical scores through the application of different rules. Overall, they gave different results on almost half of the tweets because 46% had different scores on one or both of the scales.

Table 13. SentiStrength negativity (rows) against TensiStrength stress (columns) for the full collection of 7274680 tweets matching keyword searches, after eliminating duplicates.

| Negativity\Stress | -1 | -2 | -3 | -4 | -5 |
|---|---|---|---|---|---|
| -1 | 30.9% | 1.4% | 0.4% | 0.4% | 0.1% |
| -2 | 1.8% | 28.4% | 6.2% | 0.8% | 0.1% |
| -3 | 0.3% | 7.0% | 10.0% | 1.1% | 0.1% |
| -4 | 0.1% | 0.6% | 5.1% | 3.8% | 0.2% |
| -5 | 0.0% | 0.0% | 0.3% | 0.6% | 0.4% |

Table 14. SentiStrength positivity (rows) against TensiStrength relaxation (columns) for the full collection of 7274680 tweets matching keyword searches, after eliminating duplicates.

| Positivity\Relaxation | 1 | 2 | 3 | 4 | 5 |
|---|---|---|---|---|---|
| 1 | 55.0% | 1.9% | 0.3% | 0.5% | 0.0% |
| 2 | 9.1% | 12.0% | 1.3% | 0.3% | 0.0% |
| 3 | 1.6% | 7.8% | 5.8% | 0.3% | 0.0% |
| 4 | 0.2% | 0.4% | 2.4% | 0.8% | 0.0% |
| 5 | 0.0% | 0.0% | 0.2% | 0.3% | 0.0% |

**Comparison between supervised and unsupervised versions** The supervised version of TensiStrength works slightly better than the unsupervised version, especially on the larger corpus even though it is more varied than the sub-corpora. The MAD improvement is larger for the relaxation strength detection than for stress strength detection. The improvements suggest that additional fine tuning of the term strengths is necessary and that the supervised version of TensiStrength, in general, is preferable to the unsupervised variant, when training data is available. The adjusted term weights after supervision can be retained for future applications of TensiStrength in order to take advantage of the improved performance.

**Comparison with machine learning** The generic machine learning methods exceeded the performance of TensiStrength in most contexts, with the main exception being for stress strength detection on the transport sub-corpus. A practical disadvantage of the generic machine learning approach used, however, is that its performance is partly based on identifying stressful topics in a mixed corpus rather than specific indicators of stress. This can therefore give a misleading impression of accuracy. For example, the terms with high information gain in the corpus included a number of topic-specific neutral words, such as "military" and "hotels". Presumably, for example, a tweet containing the term *military* would be much more likely to describe stress than would other tweets. The use of this term to help detect stress might not be helpful for a corpus of tweets with a narrower topic focus – either mainly about military-related issues or mainly not about military-related issues. This stress-related topic detection might also be unhelpful in contexts where it is important to ignore the topic discussed and identify explicit expressions of stress or relaxation. In support of this, the relatively strong performance of TensiStrength for stress strength detection on the transport corpus might be due to its narrower topic focus so that the machine learning approach cannot use topic information to help predict tweets in which stress is likely to occur. In contrast, it may be that some forms of transport are more likely to be associated with relaxation, such as walking, and so the topic of a tweet might be a good indicator of the likelihood that it mentions relaxation even within a collection of transport-related texts.

# Conclusions

The results show that TensiStrength is able to detect expressions of stress and relaxation in tweets with a reasonable level of accuracy compared to human coders, more accurately than a similar sentiment analysis program, but not as accurately as generic machine learning methods optimised and trained on the same data. Unsupervised TensiStrength gives a reasonable level of accuracy on a wide range of different Twitter sub-corpora, showing that it is flexible enough to work in a range of different contexts. TensiStrength can therefore be used as an off-the-shelf solution for stress and relaxation detection, although generic machine learning methods should be chosen instead for tasks where stress-related topic detection is not a concern, and particularly for sets of texts with multiple topics.

The relatively small difference between the supervised performances of SentiStrength and TensiStrength is surprising, especially given that they are different enough to disagree on nearly half of the tweets collected. This suggests that there is a potential to further improve the accuracy of TensiStrength by identifying additional ways in which stress and relaxation are expressed through emotion. Detailed investigations into how people express stress and relaxation through text may also help to suggest improvements for TensiStrength.

Although the accuracy of TensiStrength has been evaluated abstractly, pragmatic evaluations are also needed. In other words, evidence is also needed to show that embedding it in other computer systems can lead to improved outcomes. In particular, given the motivation behind TensiStrength, it would be useful to assess whether it can help to aid the performance of transport-related systems. TensiStrength also needs to be extended and tested in contexts involving specific types of stressor, such as those related to sport, lifestyle, or work commitments. Stress and relaxation are important aspects of our daily lives and software that can accurately identify it can help to enable future smart applications as well as research into the causes and impacts of stress.

# Acknowledgement

This research was supported by the European Union's Horizon 2020 research and innovation programme under grant agreement No 636160-2, the Optimum project www.optimumproject.eu.

## Appendix 1: Online resources

TensiStrength can be tested online at: http://sentistrength.wlv.ac.uk/TensiStrength.html.

The main coder instructions are available at https://figshare.com/articles/Stress_and_Relaxation_Code_Book_for_Twitter_data/3463481

Additional instructions created by the coders to enhance consistency are at: https://figshare.com/articles/Extra_instructions_to_help_ensure_coding_consistency_for_stress_and_relaxation_coding/3463484

The spreadsheet used for coding is available at: [https://figshare.com/articles/Spreadsheet_used_for_stress_and_relaxation_coding_of_tweets/3463487](https://figshare.com/articles/Spreadsheet_used_for_stress_and_relaxation_coding_of_tweets/3463487) .

The development corpus is available at [https://figshare.com/articles/Stress_and_relaxation_in_tweets_-_development_corpus/3463490](https://figshare.com/articles/Stress_and_relaxation_in_tweets_-_development_corpus/3463490)

The evaluation corpus is available at [https://figshare.com/articles/Stress_and_relaxation_in_tweets_-_evaluation_corpus/3463493](https://figshare.com/articles/Stress_and_relaxation_in_tweets_-_evaluation_corpus/3463493)

The keywords used for the Twitter searches are available at [https://figshare.com/articles/List_of_stress_and_relaxation_terms_with_strengths/3463496](https://figshare.com/articles/List_of_stress_and_relaxation_terms_with_strengths/3463496)

The Java jar of TensiStrength is available free from the author for academic research.

## Appendix 2: Twitter Stress and Relaxation Coding Instructions

The full coding instructions are online (see Appendix 1) and the key parts are reproduced below.

**STRESS**

Stress in psychology is a feeling of pressure or strain. There are two types of stress. Distress is negative stress and is often associated with a concern about something bad happening. Eustress is positive stress that may be associated with challenges that may lead to positive outcomes. Here, were are interested in identifying tweets that directly or indirectly indicate that the tweeter or someone else is experiencing a feeling of pressure or strain, and do not differentiate between distress and eustress. For example an athlete enjoying competing in a race would be coded as undergoing stress even though it is not distress.

**Stress**: Code each tweet for the degree to which it directly describes stress, worry, fear or anger. The *stress* scale is:

[no description of stress] -**1** · **-2** · **-3** · **-4** · **-5** [description of high levels of stress]

- Allocate -1 if the tweet contains no description of stress, fear, worry or anger.
- Allocate -5 if the comment describes high levels of stress, fear, worry or anger.
- Allocate a number between -2 and -4 if the comment describes stress, fear, worry or anger in some way but not high levels. Use your judgement about the exact level present.

Examples:
- Statements of stress: I am stressed at the moment; He is under a lot of pressure.
- Statements of anger, fear or worry: I was angry to get so much spam; we are worried about the election result.

**Stressors**: Code each tweet for the degree to which it describes things or situations that may trigger stress (including, but not limited to, travel worries). The *stressors* scale is:

[no description of stressors] -**1** · **-2** · **-3** · **-4** · **-5** [description of strong stressors]

- Allocate -1 if the tweet contains no description of stressors.

- Allocate -5 if the comment describes strong stressors that are likely to cause high levels of stress.
- Allocate a number between -2 and -4 if the comment describes stressors but not strong stressors. Use your judgement about the exact level of stress present.

Examples:
- Stressful situations (stressors): We are late; I am overworked; they are stuck in a traffic jam; she is in a rush; they are very ill; I have to finish the cooking before 6, I ran for the bus, I am very busy, she is in a hurry.
- Bad things that might be causes of stress: This car is terrible.

**Negative feelings**: Code each tweet for the degree to which it describes or suggests negative feelings other than stress, fear, worry or anger. The *negative* **feelings** scale is:

[no description of negative feelings] -**1** · **-2** · **-3** [strong negative feelings]
- Allocate -1 if the tweet contains no indicators of negative feelings other than stress.
- Allocate -2 if the comment contains a moderate indicator of negative feelings (ignoring stress).
- Allocate -3 if the tweet describes a strong indicator of negative feelings (ignoring stress).

Examples:
- Negative feelings or emotions except stress: I hate Mark, Tony is sad, I am depressed.
- Negative opinions but not stressors: I dislike the camera.
- Insults (normally score these as -2): Sarah is an idiot; Nigel is stupid.
- Contexts suggesting negative emotions: He is crying. We grumbled.

**The most important number is the overall stress score – the three separate categories are mainly to guide your judgements. Don't worry about the individual scores if you are not sure which one of the above applies – as long as the overall score is the same it does not matter.**

### RELAXATION

Relaxation refers to a bodily state of reduced readiness for activity, or a feeling of peacefulness or calm. Sleep and yoga are extreme examples of relaxation. Here we are concerned with identifying relaxation, the absence of stress (i.e., information that does not just contain no stress indications but contains suggestions that stress is absent, such as "Today is a good day."), as well as confidence in the ability to cope with stress (e.g., "I am going to win the race!").

**Relaxedness**: Code each tweet for the degree to which it describes a relaxed state. The *relaxedness* scale is:

[no description of a relaxed state] **1**· **2** · **3** · **4** · **5** [description of a highly relaxed state]
- Allocate 1 if the tweet contains no description of a relaxed state.
- Allocate 5 if the tweet describes a highly relaxed state.

- Allocate a number between 2 and 4 if the describes relaxation but not high levels of relaxation. Use your judgement about the exact strength of relaxation.

Examples:
- I am very relaxed; they are asleep; I feel peaceful; they are calm.

**Relaxation**: Code each tweet for the degree to which it describes relaxing activities or contexts. The *relaxation* scale is:

[no description of relaxing activities] **1 · 2 · 3 · 4 · 5** [description of highly relaxing activities]

- Allocate 1 if the tweet contains no description of relaxing activities.
- Allocate 5 if the tweet describes highly relaxing activities.
- Allocate a number between 2 and 4 if the describes relaxing activities but not highly relaxing activities. Use your judgement about the exact strength.

Examples:
- Relaxing activities: I went for a stroll; we wandered around town; she is reading a good book; we were meditating; she dozed off.
- Relaxing contexts: I am in a sauna; we have plenty of time before the train departs; it is a quiet day; some slow music is playing quietly in the background.

**Positive feelings:** Code each tweet for the degree to which it describes or suggests positive feelings, other than relaxed states. The *positive feelings* scale is:

[no indicators of positive feelings] **1 · 2 · 3** [strong indication of positive feelings]

- Allocate 1 if the tweet contains no indicators of positive feelings.
- Allocate 2 if the tweet contains indications of moderately positive feelings.
- Allocate 3 if the tweet contains indications of strong positive feelings.

Examples:
- Descriptions of positive feelings or emotions: I am happy; I feel wonderful.
- Positive opinions: The car is excellent; that is lovely.
- Things associating with positive feelings: She is smiling; They kissed; It's her birthday.

**The most important number is the overall relaxation score – the three separate categories are mainly to guide your judgements. Don't worry about the individual scores if you are not sure which one of the above applies – as long as the overall score is the same it does not matter.**